\documentclass[runningheads]{llncs}
\usepackage{graphicx}
\usepackage{comment}
\usepackage{amsmath,amssymb} 
\usepackage{color}

\usepackage[ruled]{algorithm2e}
\usepackage{color}
\usepackage{booktabs}
\usepackage{threeparttable}
\usepackage{float}
\usepackage{bm}
\usepackage{multirow}
\usepackage{caption}
\usepackage{amssymb}
\usepackage{wrapfig}
\usepackage{subcaption}
\usepackage{tabularx}
\usepackage{appendix}
\usepackage[colorlinks=true]{hyperref}
\usepackage{cleveref}

\begin{document}
\pagestyle{headings}
\mainmatter
\def\ECCVSubNumber{5004}  

\title{Accelerating CNN Training by Pruning Activation Gradients
\thanks{This work is supported in part by the National Natural Science Foundation of China (61602022), State Key Laboratory of Software Development Environment (SKLSDE-2018ZX-07), CCF-Tencent IAGR20180101 and the 111 Talent Program B16001.}
} 

\titlerunning{Accelerating CNN Training by Pruning Activation Gradients}
%
\author{Xucheng Ye\inst{1} \and
Pengcheng Dai \inst{2} \and
Junyu Luo\inst{1} \and
Xin Guo \inst{1} \and
Yingjie Qi \inst{1} \and \\
Jianlei Yang \inst{1} \and
Yiran Chen \inst{3} }
\authorrunning{Xucheng et al.}
%
\institute{SCSE, BDBC, Beihang University, Beijing, China
\and
SME, BDBC, Beihang University, Beijing, China \and
ECE, Duke University, Durham, NC, USA \\
\email{jianlei@buaa.edu.cn}
 }

\maketitle

\begin{abstract}

Sparsification is an efficient approach to accelerate CNN inference, but it is challenging to take advantage of sparsity in training procedure because the involved gradients are dynamically changed. Actually, an important observation shows that most of the activation gradients in back-propagation are very close to zero and only have a tiny impact on weight-updating. Hence, we consider pruning these very small gradients randomly to accelerate CNN training according to the statistical distribution of activation gradients. Meanwhile, we theoretically analyze the impact of pruning algorithm on the convergence. The proposed approach is evaluated on AlexNet and ResNet-\{18, 34, 50, 101, 152\} with CIFAR-\{10, 100\} and ImageNet datasets. Experimental results show that our training approach could substantially achieve up to $5.92 \times$ speedups at back-propagation stage with negligible accuracy loss.


\end{abstract}

\section{Introduction}

Convolutional Neural Networks (CNNs) have been widely applied to many tasks and various devices in recent years. However, the network structures are becoming more and more complex, making the training of CNN on large scale datasets very time consuming, especially with limited hardware resources. Some previous researches have shown that CNN training  could be finished within minutes on high performance computation platforms \cite{goyal2017accurate}\cite{you2018imagenet}\cite{jia2018highly}, but thousands of GPUs have to be utilized, which is not feasible for many scenarios. Even though there are many existing works on network compressing, most of them are focused on inference \cite{cheng2018recent}. Our work aims to reduce the training workloads efficiently, enabling large scale training on budgeted computation platforms.

The essential optimization step of CNN training is to perform Stochastic Gradient Descent (SGD) algorithm in back-propagation procedure. There are several data types involved in training dataflow: weights, weight gradients, activations, activation gradients. Back-propagation starts from computing the weight gradients with the activations and then performs weights update \cite{micikevicius2017mixed}. Among these steps, \textit{activation gradients back-propagation} and \textit{weight gradients computation} require intensive convolution operations thus dominate the total training cost. It is well known that computation cost can be reduced by skipping over zero-values. Since these two convolution steps require the activation gradients as input, improving the sparsity of activation gradients should significantly reduce the computation cost and memory footprint during back-propagation procedure.

Without loss of generality, we assume that the numerical values of activation gradient satisfy normal distributions, and a threshold ${\tau}$ can be calculated based on this hypothesis. And then \textit{stochastic pruning} is applied on the activation gradients with the threshold ${\tau}$ while the gradients are set to zero or ${\pm\tau}$ randomly.
Since the \texttt{ReLU} layers usually make the gradients distributed irregularly, we divide common networks into two categories, one is networks using \texttt{Conv-ReLU} as basic blocks such as AlexNet \cite{krizhevsky2012imagenet} and VGGNet \cite{simonyan2014very}, another is those using \texttt{Conv-BN-ReLU} structure such as ResNet \cite{he2016deep}.
Experiments show that our pruning method works for both \texttt{Conv-ReLU} structure and \texttt{Conv-BN-ReLU} structure in modern networks.
A mathematical analysis is provided to demonstrate that stochastic pruning can maintain the convergence properties of CNN training.
Additionally, our proposed training scheme is evaluated both on Intel CPU and ARM CPU platforms, which could achieve $1.71 \times \sim 3.99 \times$ and $1.79 \times \sim 5.92 \times$ speedups, respectively, when compared with no pruning utilized at back-propagation stage.
\section{Related Works} \label{gen_inst}

\paragraph{\textbf{Weight pruning}} is a well-known acceleration technique for CNN inference phase which has been widely researched and achieved outstanding advances. 
Pruning of weights can be divided into five categories  \cite{cheng2018recent}: element-level \cite{han2015deep}, vector-level \cite{mao2017exploring}, kernel-level \cite{anwar2017structured}, group-level \cite{lebedev2016fast} and filter-level pruning  \cite{luo2017thinet}\cite{he2017channel} \cite{liu2017learning}\cite{wen2017coordinating}\cite{wen2017learning}. Weight pruning focuses on raising parameters sparsity of convolutional layers.

\paragraph{\textbf{Weight gradients pruning}} is proposed for training acceleration by reducing communication cost of weight gradients exchanging in distributed learning system. 
Aji \cite{aji2017sparse} prunes $99\%$ weight gradients with the smallest absolute value by a heuristic algorithm.
According to filters’ correlationship, Prakash \cite{prakash2018repr} prunes $30\%$ filters temporarily to improve training efficiency. 

\paragraph{\textbf{Activation gradients pruning}} is another approach to reduce training cost but is rarely researched because activation gradients are generated dynamically during back-propagation. Most previous works adopt \emph{top-k} as the base algorithm for sparsification. 
For MLP training, Sun \cite{sun2017meprop} adopts \textit{min-heap} algorithm to find and retain the \emph{k} elements with the largest absolute value in the activation gradients for each layer, and discards the remaining elements to improve sparsity. Wei \cite{wei2017minimal} further applies this scheme to CNN's training, but only evaluated on LeNet. In the case of larger networks and more complex datasets, directly dropping redundant gradients will cause significant loss of learnt information. To alleviate this problem, Zhang \cite{zhang2019memorized} stores the un-propagated gradients at the last learning step in memory and adds them to the gradients before \emph{top-k} sparsification in the current iteration.
Our work can be categorized into this scope. We propose two novel algorithms to determine the pruning threshold and preserve the valuable information, respectively.


\paragraph{\textbf{Quantization}} is another common way to reduce the computational complexity and memory consumption of training.
Gupta’s work \cite{gupta2015deep} maintains the accuracy by training the model in the precision of 16-bit fixed-point number with stochastic rounding. 
DoReFaNet \cite{zhou2016dorefa} derived from AlexNet \cite{krizhevsky2012imagenet} utilizes 1-bit, 2-bit and 6-bit fixed-point number to represent weights, activations and gradients respectively, but brings visible accuracy drop. 
Park \cite{park2018value} proposed a value-aware quantization method by using low-precision on small values, which can significantly reduce memory consumption when training ResNet-152 \cite{he2016deep} and Inception-V3 \cite{szegedy2016rethinking} with $98\%$ activations quantified to $3$-bit.
Micikevicius \cite{micikevicius2017mixed} keeps an FP32 copy for weight update and adopts FP16 for computation, which is efficient for training acceleration. Our approach can be regarded as gradients sparsification, and can be also integrated with gradients quantization methods.

\section{Methodologies} \label{headings}
  \subsection{General Dataflow} \label{sec-3-1}
     The convolution (\texttt{Conv}) layer involved in each training iteration usually includes four stages: \emph{Forward}, \emph{Activation Gradients Back-propagation}, \emph{Weight Gradients Computation} and \emph{Weight Update}. To present the calculation of these stages, some definitions and notations are introduced and adopted throughout this paper:
     \begin{itemize}
     \item   $\mathbf{I}$ denotes the input of each layer at \emph{Forward} stage.
     \item    $\mathbf{O}$ denotes the output of each layer at \emph{Forward} stage.
     \item   $\mathbf{W}$ denotes the weights of \texttt{Conv} layer.
     \item   $\mathrm{d}\mathbf{I}$ denotes the gradients of I.
     \item   $\mathrm{d}\mathbf{O}$ denotes the gradients of O.
     \item   $\mathrm{d}\mathbf{W}$ denotes the gradients of W.
     \item   $\ast$ denotes the 2-D convolution.
     \item   $\eta$ denotes the learning rate.
     \item   $\mathbf{W}^+$ denotes the sequentially reversed of $\mathbf{W}$.
    \end{itemize}
    And the four training stages of \texttt{Conv} layer can be summarized as:
      \begin{itemize}
      \item \emph{Forward} ${\mathbf{O} = \mathbf{I} \ast \mathbf{W}}$ (notice that we leave out bias here)
      \item \emph{Activation Gradients Back-Propagation}(\emph{AGBP}):  $\mathrm{d}\mathbf{I}$ = $\mathbf{{W}^+}$ $\ast$ $\mathrm{d}\mathbf{O}$
      \item \emph{Weight Gradients Computation}(\emph{WGC}): $\mathrm{d}\mathbf{W}$ = $\mathrm{d}\mathbf{O}$ $\ast$ $\mathbf{I}$
      \item \emph{Weight Update}: $\mathbf{W}$ $\leftarrow$ $\mathbf{W}$ - $\eta$ $\cdot$ $\mathrm{d}\mathbf{W}$
    \end{itemize}


    We found that activation gradients involved in back-propagation stage are almost full of \emph{very small values} that are extremely close to zero.
    It is reasonable to assume that pruning those extremely small values has little effect on weight update stage.
    Meanwhile, existing works show that pruning redundant elements in convolution calculations can effectively reduce arithmetic complexity.
    Therefore, we make a hypothesis that the involved \texttt{Conv} layers computations in training can be accelerated substantially by pruning activation gradients.

    \begin{figure}[t]
    \centering
    \begin{minipage}[c]{0.45\linewidth}
        \centering
        \includegraphics[width=0.75\textwidth]{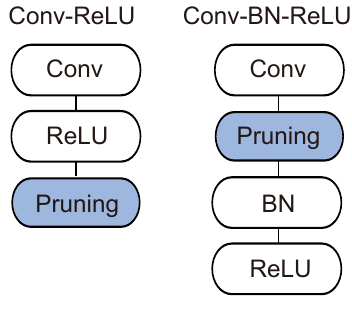}
        \caption{Pruning stages involved for two typical structures: \texttt{Conv-ReLU} and \texttt{Conv-BN-ReLU}.}
        \label{fig-position}
      \end{minipage}
      \hfill
      \begin{minipage}[c]{0.46\linewidth}
        \centering
        \includegraphics[width=\textwidth]{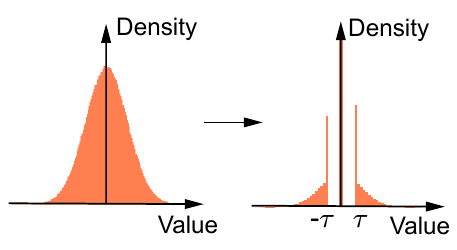}
        \caption{Effect of \textit{stochastic pruning}, where $\tau$ is the pruning threshold.}
        \label{fig-pruning-effect}
      \end{minipage}
    \end{figure}

\subsection{Sparsification Algorithms}\label{sec-3-2}

   \paragraph{\textbf{Distribution Based Threshold Determination (DBTD).}}
   The most important concern of pruning is to determine which elements should be selected for discarding.
   Previous works \cite{sun2017meprop} use min-heap algorithm to select which elements going to be pruned. However, they will introduce inevitable overhead significantly when implemented on heterogeneous platforms such as FPGA or ASIC. Hence, we propose a new threshold determination method with less time complexity and more hardware compatibility.

   Firstly, we analyze the distribution of activation gradients for two typical structures of modern CNN models, as shown in Fig. \ref{fig-position}.
   For \texttt{Conv-ReLU} structure, where a \texttt{Conv} layer is followed by a \texttt{ReLU} layer, output activation gradients $\mathrm{d}\mathbf{O}$ are sparse, but subject to an irregular distribution.
   On the other hand, the input activation gradients $\mathrm{d}\mathbf{I}$, which will be propagated to the previous layer, is almost full of non-zero values.
   Statistics show that the probability distribution of $\mathrm{d}\mathbf{I}$ is symmetrical around zero and its probability density function decreases with the increment of absolute value $\left|  \mathrm{d}\mathbf{I} {\left( { \cdot } \right)} \right|$.
   For \texttt{Conv-BN-ReLU} structure, a \texttt{BN} layer is located between \texttt{Conv} and \texttt{ReLU} layer, and  $\mathrm{d}\mathbf{O}$ subjects to the similar distribution of $\mathrm{d}\mathbf{I}$.
   With the same hypothesis \cite{wen2017terngrad}, these gradients are assumed to subject to a normal distribution with mean value $0$ and variance ${{\sigma^2}}$.

   For \texttt{Conv-ReLU} structure, $\mathrm{d}\mathbf{O}$ can inherit the sparsity from $\mathrm{d}\mathbf{I}$ of last \texttt{Conv} layer because \texttt{ReLU} layer will not reduce the sparsity.
   Thus $\mathrm{d}\mathbf{I}$ can be treated as pruning target $g$ in \texttt{Conv-ReLU} structure.
   For \texttt{Conv-BN-ReLU} structure, $\mathrm{d}\mathbf{O}$ is considered as pruning target $g$.
   In this way, the distribution of $g$ in both situations could be unified to normal distribution.
   Supposing that the scale of $g$ is $n$, we calculate the mean value of the absolute values from gradient data $g$, and the expectation of it is:
    \begin{equation}
        E\left(\frac{1}{n} \sum\limits_{i = 1}^n {\left| {{g_i}} \right|} \right)
            = \frac{{n}}{{\sqrt {2\pi {\sigma ^2}} }}\int {\left| x \right|\exp \left\{ { - \frac{{{x^2}}}{{2{\sigma ^2}}}} \right\}dx}
            = \sqrt {\frac{2}{\pi }} n\sigma .
    \end{equation}
    Let
    \begin{equation}
        \hat \sigma  = \frac{1}{n}\sqrt {\frac{2}{\pi }} \sum\limits_{i = 1}^n {|{g_i}|} ,
    \end{equation}
    then
    \begin{equation}
        E(\hat \sigma ) = E\left(\frac{1}{n}\sqrt {\frac{2}{\pi }} \sum\limits_{i = 1}^n {|{g_i}|} \right) = \sigma .
    \end{equation}
    Clearly, ${\hat \sigma}$ is an unbiased estimator of parameter $\sigma$.

    Here we adopt the mean value of the absolute values because the computational overhead is acceptable.
    Base on the assumption,  we can compute the threshold ${\tau}$ with the cumulative distribution function of the standard normal distribution ${\Phi}$, target pruning rate $p$ and ${\hat \sigma}$ by:
    \begin{equation}
        \tau  = \Phi^{-1} \left( {\frac{1 - p}{2}} \right)\hat \sigma .
    \end{equation}

    \paragraph{\textbf{Stochastic Pruning.}}
    Pruning a few gradients with small values has little impact on weights update. However, once all of these small gradients are set to $0$, the distribution of activation gradients will be affected significantly, which will influence the weights update and cause severe accuracy loss.
    Inspired by \emph{Stochastic Rounding} in \cite{gupta2015deep}, we adopt stochastic pruning to solve this problem.

    Stochastic pruning treats gradients as an one-dimensional vector $g$ with length $n$, and all the components whose absolute value is smaller than the threshold $\tau$ will be pruned. The algorithm details are demonstrated in Algorithm \ref{algo:1}. The effect of stochastic pruning on gradient distribution is illustrated in Fig. \ref{fig-pruning-effect}.

    \begin{algorithm}[H] \label{algo:1}
      \caption{Stochastic Pruning}
      \KwIn{original activation gradients ${g}$, threshold ${\tau}$}
      \KwOut{sparse activation gradients ${\hat g}$ }
      \For{$i=1;i \le n;i=i+1$}
      {
          \If{${|g_i| < \tau}$}
          {
            Generate a random number $r \in \left[ {0,1} \right]$ \;
            \eIf{$|g_i| > r \tau $}
            {
              ${\hat g_i} = ({g_i} > 0) $ ? ${\tau}$ : $({-\tau})$ \;
            }
            {
              ${\hat g_i}$ = 0 \;
            }
          }

    }
    \end{algorithm}

    Stochastic pruning could maintain the mathematical expectation of the gradients distribution while completing the pruning. Mathematical analysis in Section \ref{sec:convergence-analysis} will show that such a gradients sparsification method for CNN training does not affect its convergence.

    In summary, compared with existing works, our scheme has two advantages:
    \begin{enumerate}
    \renewcommand{\labelenumi}{(\theenumi)}
         \item \textit{Lower runtime cost}: the arithmetic complexity of \textit{DBTD} is $\mathcal{O}\left( n \right)$, less than \texttt{top-k} which is at least $\mathcal{O}\left( n \log k\right)$, where $k$ stands for the number of reversed elements. Meanwhile, \textit{DBTD} is more hardware friendly and easier to be implemented on heterogeneous platform because it does not require frequent comparison operations.
         \item \textit{Lower memory footprint}: our \textit{Stochastic Pruning} approach could preserve the convergence rate and does not require any extra memory consumption. In contrast, \cite{zhang2019memorized} needs to store the un-propagated gradients of the last training steps, which is more memory consuming.
    \end{enumerate}

\newtheorem{assumption}{Assumption}

\section{Convergence Analysis \label{sec:convergence-analysis}}
  The convergence rate of our proposed stochastic pruning is analyzed in this section.
  \textit{Please note that it is not a rigorous mathematical proof, but just provide some intuition on why the gradients pruning method works.}
  We expect that our training method with stochastic pruning has similar convergence rate with origin training process under the GOGA (General Online Gradient Algorithm) framework \cite{bottou1998online}.

  In \cite{bottou1998online}, L.~Bottou considers a learning problem as follows: suppose that there is an unknown distribution $P(z)$ and can only get a batch of samples $z_t$ each iteration, where $t$ denotes iteration times. The goal of training is to find the optimal parameters $w$ which minimize the loss function $\mathrm{Q}(z, w)$. 
  For convenience, we define the cost function as:
  \begin{equation}
    \mathrm{C}(w) \triangleq \mathbf{E_z} \mathrm{Q}(z, w)
    \triangleq \int \mathrm{Q}(z, w)\,\mathrm{d}P(z) .
  \end{equation}
  And the involved update rule for the online learning system is formulated as: 
  \begin{equation}
    w_{t+1} = w_t - \gamma_t \mathrm{H}\left( z_t, w_t \right),
  \end{equation}
  where $\gamma_t$ is the learning rate, $\mathrm{H}\left( \cdot \right)$ is the update function. It will finally converge as long as the following assumptions are satisfied.
  \begin{assumption} \label{assu:1}
    The cost function $\mathrm{C}(w_t)$ has a single global minimum $w^*$ and satisfies the condition that 
    \begin{equation} \label{eq:assu1}
      \forall \varepsilon,\quad \inf_{(w - w^*)^2 > \varepsilon} (w - w^*) \nabla_w \mathrm{C}(w) > 0 .
    \end{equation}
  \end{assumption}
  
  \begin{assumption} \label{assu:2}
    Learning rate $\gamma_t$ fulfills that 
    \begin{equation} \label{eq:assu2}
      \sum_{t=1}^\infty \gamma_t = \infty, \quad\quad\quad and \quad\quad\quad \sum_{t=1}^\infty \gamma_t^2 < \infty .
    \end{equation}
  \end{assumption}
  
  \begin{assumption} \label{assu:3}
    For each iteration, the update function $\mathrm{H}(z_t, w_t)$ meets that
    \begin{equation} \label{eq:assu3}
      \mathbf{E}  \left[ \mathrm{H}(z, w) \right] = \nabla_w \mathrm{C}(w) ,
    \end{equation}
    and 
    \begin{equation} \label{eq:assu4}
      \mathbf{E} \left[ \mathrm{H}(z, w)^2 \right] \le \alpha + \beta (w - w^*)^2 ,
    \end{equation}
  where $\alpha$ and $\beta$ are finite constants, the update function $\mathrm{H}(z, w)$ consists of the calculated gradients by back-propagation algorithm.
  \end{assumption}

  The only difference between our proposed algorithm and \cite{bottou1998online} is the update function $\mathrm{H}(z, w)$. For original algorithm, the update function $\mathrm{H}(z_t, w_t)$ satisfies:
  \begin{equation}
    \mathbf{E} \left[ \mathrm{H}(z, w) \right] = \nabla_w \mathrm{C}(w) .
  \end{equation}
  In proposed algorithm, a gradients pruning method is applied on the update function, denoted as $\mathrm{\hat H}(z, w)$. 
  In this case, if we assume original back-propagation algorithm meets all the assumptions, the proposed algorithm also satisfies \textit{Assumption} \ref{assu:1} and \ref{assu:2}. If \textit{Assumption} \ref{assu:3} can be also held by the proposed algorithm, we can say that both algorithms have similar convergence.
  For convenience, their corresponding gradients are denoted as $G \triangleq \mathrm{H}(z, w)$ and $\hat G \triangleq \mathrm{\hat H}(z, w)$.
  
  \textbf{In the following we will first prove that though $\hat G \ne G$, the expectations of them are the same. What's more, we expect the extra noise introduced by gradient pruning is not significant enough to violate \textit{Assumption} \ref{assu:3}.} More precisely, the following equations should be held:
  \begin{align}
    \mathbf{E} \left[ \hat G \right] &= \mathbf{E} \left[ G \right] \label{eq:t1-l1} , \\
    \mathbf{E} \left[ {\hat G}^2 \right] &\le \alpha + \beta \mathbf{E} \left[ G^2 \right] \label{eq:t2-l1} .
  \end{align}

  To discuss \textit{Assumption} \ref{assu:3}, we first give a lemma:
  \begin{lemma} \label{lemma:1}
  For a stochastic variable $x$, we get another stochastic variable $y$ by applying \textit{Algorithm}~\ref{algo:1} to $x$ with threshold $\tau$, which means
  \begin{equation}
      y = \mathrm{Prune}(x) = \left\{
      \begin{aligned}
        & x \qquad\text{i.f.f. } |x| \ge \tau \\
        & 0 \qquad\text{with probability $p=\frac{\tau-x}{\tau}$~ i.f.f. } |x| < \tau \\
        & \tau \qquad\text{with probability $p=\frac{x}{\tau}$~ i.f.f. } |x| < \tau 
      \end{aligned} \right .
  \end{equation}
  Then $y$ satisfies
  \begin{align}
        \mathrm{E}[y] = \mathrm{E}\left[\mathrm{Prune}(x)\right] = \mathrm{E}[x] , \label{eq:lemma-ex} \\
        \mathrm{E}[y^2] = \mathrm{E}\left[\mathrm{Prune}(x)^2\right] \le \tau^2 + \mathrm{E}[x^2] . \label{eq:lemma-var} 
  \end{align}
  \end{lemma}
  
  Then we can discuss the expectation and variance of gradients $\hat G$.
  
\subsection{Expectation of Gradients}
  \textit{Lemma}~\ref{lemma:1} means that gradients pruning will not affect the expectation of activation gradients, which can be utilized to prove Eq. (\ref{eq:t1-l1}). 
  Let $G$ represent the gradients of the whole network parameters with $N$ layers. Thus we can split it into layer-wise gradients:
  \begin{equation}
    G = ( G_1, G_2, \cdots, G_l, \cdots, G_N ) 
  \end{equation}
  where $G_l$ represents the gradients of $l$-th layer weights. Let $GO_l$ represents the activation gradients for $l$-th layer, we have:
  \begin{equation}
    GO_l = F_1(GO_{l+1}, \omega) , \quad\quad\quad and \quad\quad\quad    G_l = F_2(GO_l)
  \end{equation}
  where $F_1$ and $F_2$ represents the back-propagation operation for $l$-th layer.

  The same thing can be done for $\hat G$ which means:
  \begin{align}
    \hat G &= ( \hat G_1, \hat G_2, \cdots, \hat G_l, \cdots, \hat G_N ) , \\
    \hat {GO}_l &= \mathrm{Prune}\left[ F_1({\hat {GO}}_{l+1}, \omega) \right] , \\
    \hat G_l &= F_2\left( \hat {GO}_l \right) .
  \end{align}

  To prove Eq. (\ref{eq:t1-l1}), we only need to prove that for each $l$,
  \begin{align}
    \mathbf{E} \left[ \hat G_l \right] &= \mathbf{E} \left[ G_l \right] \label{eq:t1-l2} ,\\
    \mathbf{E} \left[ \hat {GO}_l \right] &= \mathbf{E} \left[ GO_l \right] \label{eq:t2-l2} .
  \end{align}
  Note that Eq. (\ref{eq:t2-l2}) is already held for the last layer. Because the last layer is the start of back-propagation and the proposed algorithm is the same with original algorithm before the last layer's gradients $G$ are calculated, we only need to prove that:
  \begin{align}
    \mathbf{E} \left[ G_l \right] &= F_1 \left( \mathbf{E} \left[ GO_l \right] \right) \label{eq:t1-l3} \\
    \mathbf{E} \left[ {GO}_l \right] &= F_2 \left( \mathbf{E} \left[ GO_{l+1} \right] \right) \label{eq:t2-l3}
  \end{align}

  Because if Eq. (\ref{eq:t1-l3}) and Eq. (\ref{eq:t2-l3}) are held, we can prove Eq. (\ref{eq:assu3}) by using \textit{Lemma}~\ref{lemma:1}. 
  \begin{proof}
    Assume Eq. (\ref{eq:t2-l2}) is satisfied for $(l+1)$-th layer. Then for $l$-th layer
    \begin{align}
      \mathrm{E}[\hat G_l] &= F_1 \left( \mathbf{E} \left[ \hat {GO}_l \right] \right) \label{pf:1}\\
        &= F_1 \left( F_2 \left( \mathbf{E} \left[ \mathrm{Prune} \left( \hat {GO}_{l+1} \right) \right] \right) \right) \label{pf:2}\\
        &= F_1 \left( F_2 \left( \mathbf{E} \left[ \hat {GO}_{l+1} \right] \right) \right) \label{pf:3}\\
        &= F_1 \left( F_2 \left( \mathbf{E} \left[ {GO}_{l+1} \right] \right) \right) \label{pf:4}\\
        &= F_1 \left( \mathbf{E} \left[ {GO}_l \right] \right) \label{pf:5}\\
        &= \mathrm{E}[G_l] \label{pf:6}
    \end{align}
    The equality of Eq. (\ref{pf:1}) and Eq. (\ref{pf:6}) could be guaranteed by Eq. (\ref{eq:t1-l3}).
    Eq. (\ref{pf:2}) and Eq. (\ref{pf:5}) is true because of Eq. (\ref{eq:t2-l3}).
    Eq. (\ref{pf:3}) is right due to \textit{Lemma}~\ref{lemma:1}.
    Since the assumption Eq. (\ref{eq:t2-l2}) is true for the last layer, then for all $l$, Eq. (\ref{eq:assu3}) is right.
  \end{proof}

  As for Eq. (\ref{eq:t1-l3}) and Eq. (\ref{eq:t2-l3}) is true because they are linear operation except \texttt{ReLU} in the case of CNN and the back-propagation of \texttt{ReLU} can exchange with expectation. Here we denote the back-propagation of \texttt{ReLU} as \texttt{ReLU'}.
  \texttt{ReLU'} will set the operand to zero or hold its value. For the former one, 
  \[
    \mathbf{E} \left[ \texttt{ReLU'}(x) \right] = 0 = \texttt{ReLU'} \left(\mathbf{E} \left[ x \right] \right) .
  \]
  For the latter one, 
  \[
    \mathbf{E} \left[ \texttt{ReLU'}(x) \right] = \mathbf{E} \left[ x \right] = \texttt{ReLU'} \left(\mathbf{E} \left[ x \right] \right) .
  \]

  Thus we prove that the expectation of gradients in the proposed algorithm is the same with the original algorithm.
  
\subsection{Variance of Gradients}

  It is difficult to prove that Eq. (\ref{eq:assu4}) can be also satisfied in the proposed gradient pruning algorithm. However, we can give some intuition that this may be right if original training method meets this condition. 
  Eq. (\ref{eq:assu4}) tell us that, to guarantee the convergence during stochastic gradient descend, variance of gradients in each step should not be too large. The proposed gradient pruning method will indeed bring extra noise to the gradients. But we believe the extra noise is not significant enough to violate Eq. (\ref{eq:assu4}).
  The extra gradients noise is determined by two factors. First is the noise generated by pruning method. Second is the propagation of the pruning noise in the following back-propagation process. 
  
  From Eq. (\ref{eq:lemma-var}) we can tell that the variance of the pruned gradients will only increase by a constant number relating to threshold $\tau$. This will certainly obey the condition in Eq. (\ref{eq:assu4}). What's more, the noise is then propagating through \texttt{Conv} and \texttt{ReLU} layers, whose operation is either linear or sublinear. Thus we can expect that the increase of variance will still be quadratic, which satisfies Eq. (\ref{eq:assu4}). In this way, we can say that the proposed pruning algorithm has almost the same convergence with the original algorithm under the GOGA framework.
   
\section{Implementation}

\subsection{Accuracy Evaluation}

%
%
%
PyTorch \cite{paszke2017automatic} framework is utilized to estimate the impact on accuracy for our gradient pruning method. The straight-through estimator (STE) is adopted in our implementation. We introduce an extra \texttt{Pruning} layer for different \texttt{Conv} block as shown in Fig. \ref{fig-position}. As mentioned above, the input and output of this layer can be denoted as $\mathbf{I}$ and $\mathbf{O}$. The essence of this \texttt{Pruning} layer is a STE which can be defined as below:
\[
    \textbf{Forward: } {\mathbf{O} = \mathbf{I}}
\]
\[
    \textbf{Backward: }{\mathrm{d}\mathbf{I} = \textit{Stochastic\_Pruning} \left( \mathrm{d}\mathbf{O}, \textit{DBTD}(\mathrm{d}\mathbf{O}, p) \right)}
\]

\subsection{Speedup Evaluation}

To estimate the acceleration effect of our algorithm, we modify the backward \texttt{Conv} layers in Caffe \cite{jia2014caffe} framework, which is widely used in deep learning deployment. 
As mentioned in Section \ref{sec-3-1}, two main steps of training stage: \emph{AGBP}, \emph{WGC} are all based on convolution. 
Most modern deep learning frameworks including Caffe convert convolution into matrix multiplication by applying the combination of \texttt{im2col} and \texttt{col2im} functions, where \texttt{im2col} turns a 3-D feature map tensor into a 2-D matrix for exploiting data reuse, and \texttt{col2im} is the inverse function of \texttt{im2col}. Hence, our training acceleration with sparse activation gradients can be accomplished by replacing the original matrix multiplication with sparse matrix multiplication.

With our proposed algorithm, the activation gradients $\mathrm{d}\mathbf{O}$ can be fairly sparse. However, weight $\mathbf{W}$ and activation $\mathbf{O}$ are completely dense. 
We found that \texttt{dense $\times$ sparse} matrix multiplication is required for \emph{AGBP} step. However, the existing BLAS library such as Intel MKL only supports \texttt{ sparse $\times$ dense} multiplication. To solve this problem, we turn to compute the transpose of $\mathrm{d}\mathbf{I}$ according to the basic property of matrix multiplication $\left({AB}\right)^T$ = $B^TA^T$, where both $A$ and $B$ are matrices.
 
To reduce the computation cost, we modify the original \texttt{im2col} and \texttt{col2im} functions to \texttt{im2col\_trans} and \texttt{col2im\_trans} so that we can get transposed matrix directly after calling these functions. Since plenty of runtime can be saved by using \texttt{sparse $\times$ dense} multiplication, we can also achieve relatively high speedup in the overall back-propagation process, though transpose functions will cost extra runtime. The modified procedure can be summarized as: 
\[
    \emph{AGBP}:\mathrm{d}\mathbf{I} =  \texttt{col2im\_trans}\left( \texttt{sdmm}\left({\texttt{im2col\_trans}\left( {\mathrm{d}\mathbf{O}}\right), \texttt{transpose}\left(\mathbf{W}\right)} \right)\right)
\]
\[
    \emph{WGC}:\mathrm{d}\mathbf{W} = {\texttt{sdmm}\left(
    \mathrm{d}\mathbf{O}, {\texttt{im2col}\left(\mathbf{I}\right)}
    \right)}
\]
Here $\texttt{sdmm}$ denotes the general \texttt{sparse  $\times$  dense} matrix multiplication.


%
%
%
%
%
%
%
%
%
%
%
%
%
%
\section{Experimental Results}
  In this section, experiments are conducted to demonstrate that the proposed approach could reduce the training cost significantly with a negligible model accuracy loss.

  \subsection{Datasets and Models}
    Three datasets are utilized including CIFAR-10, CIFAR-100 \cite{krizhevsky2009learning} and ImageNet \cite{deng2009imagenet}. AlexNet \cite{krizhevsky2012imagenet} and ResNet \cite{he2016deep} are evaluated while ResNet include Res-\{18, 34, 50, 101, 152\}.
    The last layer size of each model is changed in order to adapt them on CIFAR datasets. Additionally for AlexNet, the kernels in first two convolution layers are set as $3 \times 3$ with $\text{padding} = 2$ and $\text{stride} = 1$. For FC-1 and FC-2 layers in AlexNet, they are also resized to $4096 \times 2048$ and $2048 \times 2048$, respectively.
    For ResNet, kernels in first layer are replaced by $3 \times 3$ kernels with $\text{padding} = 1$ and $\text{stride} = 1$. Meanwhile, the pooling layer before FC-1 in ResNet is set to Average-Pooling with the size of $4 \times 4$.

  \subsection{Training Settings}
    All the $6$ models mentioned above are trained for $300$ epochs on CIFAR-\{10, 100\} datasets. While for ImageNet, AlexNet, ResNet-\{18, 34, 50\} are only trained for $180$ epochs due to our limited computing resources.

    The Momentum SGD is adopted for all training with $\text{momentum} = 0.9$ and $\text{weight~decay} = 5\times 10^{-4}$. Learning rate \texttt{lr} is set to $0.05$ for AlexNet and $0.1$ for the others.
    \texttt{lr-decay} is set to $0.1/100$ for CIFAR-\{10, 100\} and $0.1/45$ for ImageNet.

    \begin{table}[h]
      \centering
      \begin{threeparttable}
        \caption{Evaluation results on CIFAR-10, where \texttt{acc}\% means the training accuracy and $\rho_{nnz}$ means the average density of non-zeros.} \label{tab:Cifar10}
        \begin{tabular}{c|cc|cc|cc|cc|cc}
          \specialrule{0.8pt}{0pt}{0pt}
          \multirow{2}{*}{Model}&
          \multicolumn{2}{c|}{Baseline}&
          \multicolumn{2}{c|}{$p=70\%$}&\multicolumn{2}{c|}{$p=80\%$}&\multicolumn{2}{c|}{$p=90\%$}&\multicolumn{2}{c}{$p=99\%$} \\
          \cline{2-11}
          & \texttt{acc}\% & $\rho_{nnz}$ & \texttt{acc}\% & $\rho_{nnz}$ & \texttt{acc}\% & $\rho_{nnz}$ & \texttt{acc}\% & $\rho_{nnz}$ & \texttt{acc}\% & $\rho_{nnz}$   \\
          \hline
          AlexNet&90.50&0.09&90.34&0.01&\textbf{90.55}&0.01&90.31&0.01&89.66&0.01 \\
          ResNet-18&95.04&1&\textbf{95.23}&0.24&95.04&0.22&94.91&0.20&95.18&0.16 \\
          ResNet-34&94.90&1&95.13&0.24&95.09&0.21&\textbf{95.16}&0.19&95.02&0.15 \\
          ResNet-50&94.94&1&\textbf{95.36}&0.22&95.13&0.20&95.01&0.17&95.28&0.14 \\
          ResNet-101&95.60&1&\textbf{95.61}&0.24&95.48&0.22&95.60&0.19&94.77&0.12 \\
          ResNet-152&\textbf{95.70}&1&95.13&0.18&95.58&0.18&95.45&0.16&93.84&0.08 \\

          \specialrule{0.8pt}{0pt}{0pt}
        \end{tabular}
      \end{threeparttable}
    \end{table}

    \begin{table}[h]
      \centering
      \begin{threeparttable}
        \caption{Evaluation results on CIFAR-100, where \texttt{acc}\% means the training accuracy and $\rho_{nnz}$ means the average density of non-zeros.} \label{tab:Cifar100}
        \begin{tabular}{c|cc|cc|cc|cc|cc}
          \specialrule{0.8pt}{0pt}{0pt}
          \multirow{2}{*}{Model}&
          \multicolumn{2}{c|}{Baseline}&
          \multicolumn{2}{c|}{$p=70\%$}&\multicolumn{2}{c|}{$p=80\%$}&\multicolumn{2}{c|}{$p=90\%$}&\multicolumn{2}{c}{$p=99\%$} \\
          \cline{2-11}
          & \texttt{acc}\% & $\rho_{nnz}$ & \texttt{acc}\% & $\rho_{nnz}$ & \texttt{acc}\% & $\rho_{nnz}$ & \texttt{acc}\% & $\rho_{nnz}$ & \texttt{acc}\% & $\rho_{nnz}$   \\
          \hline
          AlexNet&67.61&0.10&67.49&0.03&\textbf{68.13}&0.03&67.99&0.03&67.93&0.02 \\
          ResNet-18&76.47&1&76.89&0.27&\textbf{77.16}&0.25&76.44&0.23&76.66&0.19 \\
          ResNet-34&77.51&1&77.72&0.24&\textbf{78.04}&0.22&77.84&0.20&77.40&0.17 \\
          ResNet-50&77.74&1&78.83&0.25&78.27&0.22&\textbf{78.92}&0.20&78.52&0.16 \\
          ResNet-101&\textbf{79.70}&1&78.22&0.23&79.10&0.21&79.08&0.19&77.13&0.13 \\
          ResNet-152&79.25&1&\textbf{80.51}&0.22&79.42&0.19&79.76&0.18&76.40&0.10 \\

          \specialrule{0.8pt}{0pt}{0pt}
        \end{tabular}
      \end{threeparttable}
    \end{table}

    \begin{table}[h]
      \centering
      \caption{Evaluation results on ImageNet, where \texttt{acc}\% means the training accuracy and $\rho_{nnz}$ means the average density of non-zeros.} \label{tab:Imagenet}
      \begin{threeparttable}
        \begin{tabular}{c|cc|cc|cc|cc|cc}
          \specialrule{0.8pt}{0pt}{0pt}
          \multirow{2}{*}{Model}&
          \multicolumn{2}{c|}{Baseline}&
          \multicolumn{2}{c|}{$p=70\%$}&\multicolumn{2}{c|}{$p=80\%$}&\multicolumn{2}{c|}{$p=90\%$}&\multicolumn{2}{c}{$p=99\%$} \\
          \cline{2-11}
          & \texttt{acc}\% & $\rho_{nnz}$ & \texttt{acc}\% & $\rho_{nnz}$ & \texttt{acc}\% & $\rho_{nnz}$ & \texttt{acc}\% & $\rho_{nnz}$ & \texttt{acc}\% & $\rho_{nnz}$   \\
          \hline
          AlexNet&56.38&0.07&\textbf{57.10}&0.05&56.84&0.04&55.38&0.04&39.58&0.02 \\
          ResNet-18&68.73&1&\textbf{69.02}&0.34&68.85&0.33&68.66&0.31&68.74&0.28 \\
          ResNet-34&\textbf{72.93}&1&72.92&0.35&72.86&0.33&72.74&0.30&72.42&0.30 \\
          \specialrule{0.8pt}{0pt}{0pt}
        \end{tabular}
      \end{threeparttable}
    \end{table}

    \begin{figure}[h]
      \begin{subfigure}{0.43\linewidth}
        \includegraphics[width=\textwidth]{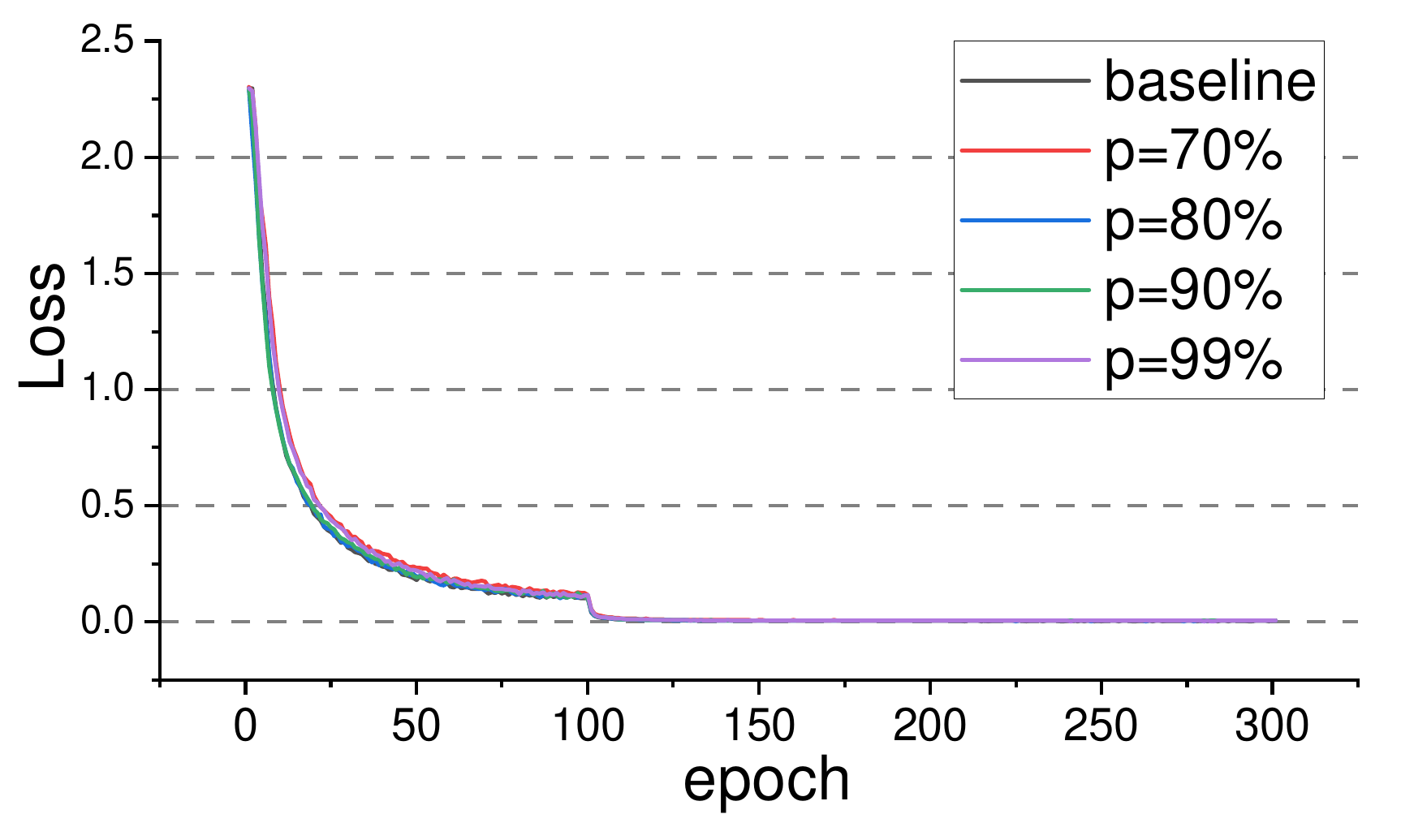}
        \caption{AlexNet on CIFAR-10.} \label{fig:loss:alexnet-cifar-10}
      \end{subfigure}
      \hfill
      \begin{subfigure}{0.43\linewidth}
        \includegraphics[width=\textwidth]{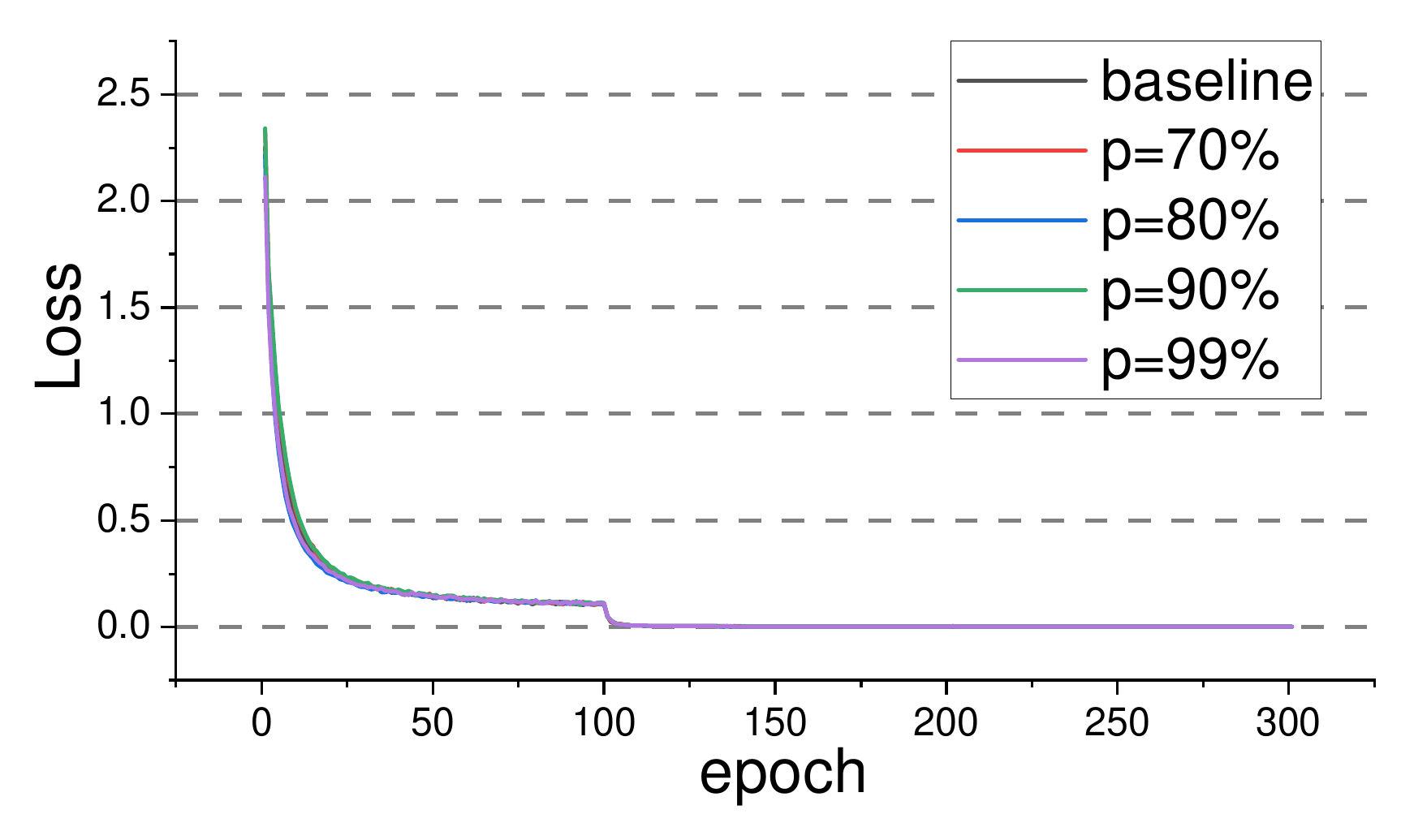}
        \caption{ResNet-18 on CIFAR-10.} \label{fig:loss:resnet-cifar-10}
      \end{subfigure}

         \par
      \begin{subfigure}{0.43\linewidth}
        \includegraphics[width=\textwidth]{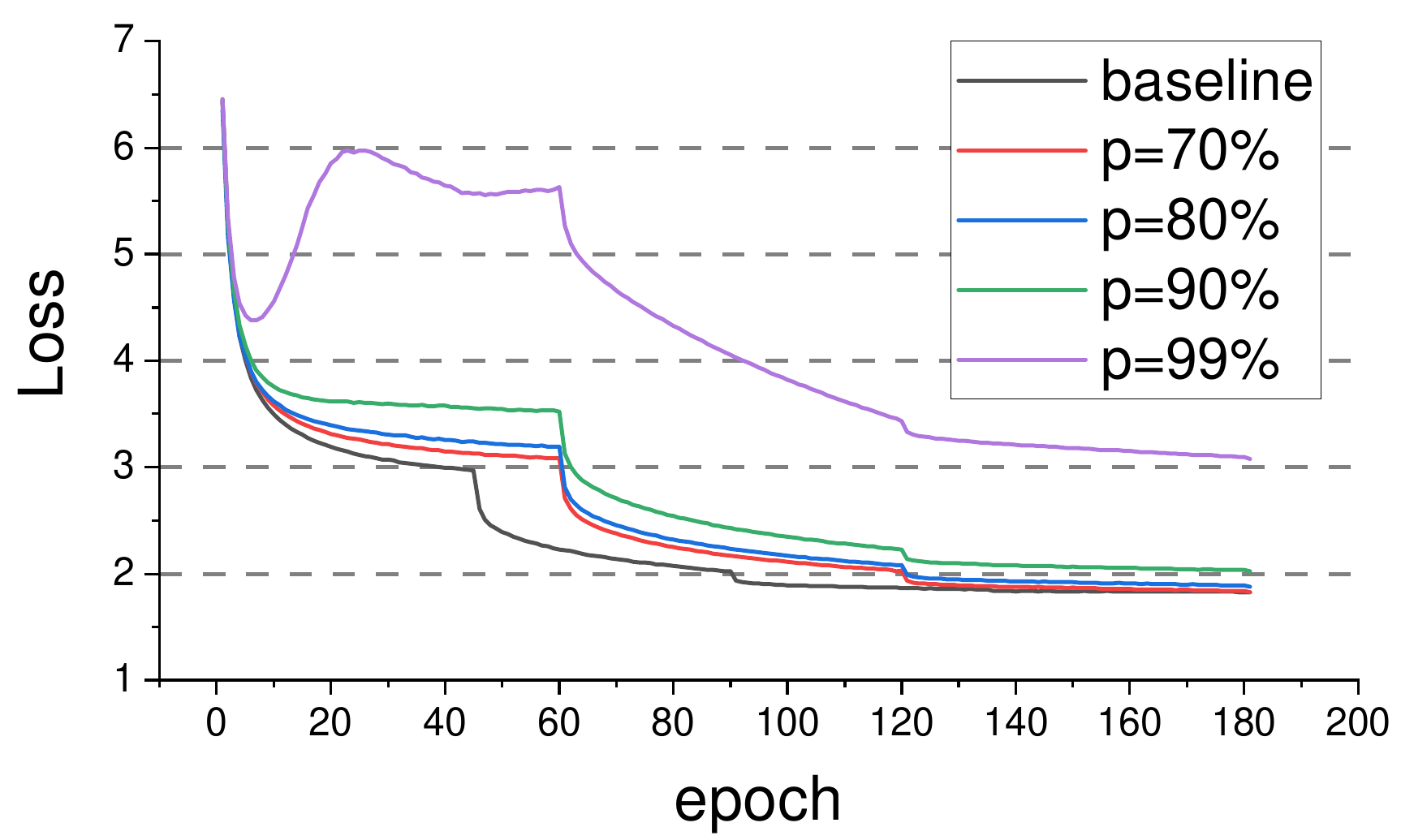}
        \caption{AlexNet on ImageNet.} \label{fig:loss:alexnet-imagenet}
      \end{subfigure}
      \hfill
      \begin{subfigure}{0.43\linewidth}
        \includegraphics[width=\textwidth]{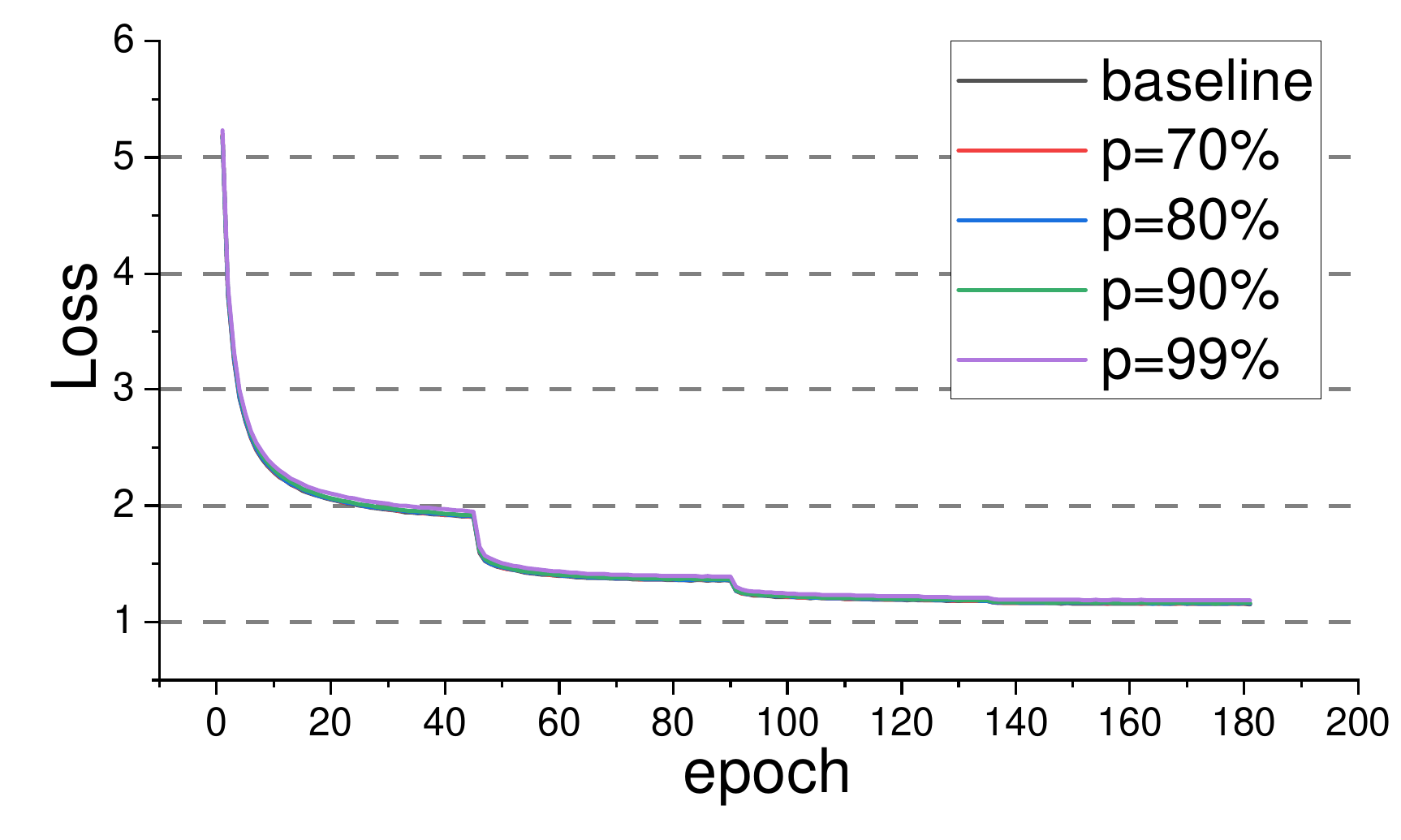}
        \caption{ResNet-18 on ImageNet.}  \label{fig:loss:resnet-imagenet}
      \end{subfigure}
        \caption{Training loss of AlexNet/ResNet on CIFAR-10 and ImageNet.}
      \label{fig:loss}

    \end{figure}

  \subsection{Results and Discussions}
    We set the target pruning rate $p$ defined in Section \ref{sec-3-2} varying from $70\%$, $80\%$, $90\%$ to $99\%$ for comparison with the baseline.
    All the training are run directly without any fine-tuning.

    \paragraph{\textbf{Accuracy Analysis.}}
    From Table \ref{tab:Cifar10}, Table \ref{tab:Cifar100} and Table  \ref{tab:Imagenet} we find that there is no obvious accuracy lost for most situations. And even for ResNet-50 on CIFAR-100, there is $1\%$ accuracy improvement. But for AlexNet on ImageNet, there is a significant accuracy loss when using very aggressive pruning policy like $p = 99\%$.
    In summary, the accuracy loss is almost negligible when a non-aggressive policy is adopted for gradients pruning.

    \paragraph{\textbf{Gradients Sparsity.}}
    The gradients density illustrated in Table \ref{tab:Cifar10}, Table \ref{tab:Cifar100}, Table \ref{tab:Imagenet} has shown the ratio of non-zero gradients over all gradients, which is related to the amount of calculations.
    Notice that the output of \emph{DBTD} is the estimation of pruning threshold, so the actual sparsity of each \texttt{Conv} layer's activation gradients will be different, and $\rho_{nnz}$ is calculated by dividing the number of non-zero activation gradients by the number of all gradients for all \texttt{Conv} layers.

    Although the basic block of AlexNet is \texttt{Conv-ReLU} whose activation gradients are relatively sparse, our method could still reduce the gradients density for about $5\times \sim 10\times$ on CIFAR-\{10, 100\} and $3\times \sim 5\times$ on ImageNet. While it comes to ResNet, whose basic block is \texttt{Conv-BN-ReLU} and activation gradients are naturally fully dense, our method could reduce the gradients density to $10\% \sim 30\%$. In addition, the deeper networks could obtain a relative lower gradients density, which means that it works better for complicated networks.

    \paragraph{\textbf{Convergence Rate.}}
    The training loss is also displayed Fig. \ref{fig:loss} for AlexNet, ResNet-18 on CIFAR-10 and ImageNet datasets. Fig. \ref{fig:loss:resnet-cifar-10} and Fig. \ref{fig:loss:resnet-imagenet} show that ResNet-18 is very robust for gradients pruning. For AlexNet, the gradients pruning could be still robust on CIFAR-10. However, Fig. \ref{fig:loss:resnet-imagenet} confirms that sparsification with a larger $p$ will impact the convergence rate. In conclusion, our pruning method doesn't have significant effect on the convergence rate in most cases. This conclusion accords with the our convergence analysis on Section \ref{sec:convergence-analysis}.

    \begin{figure}[ht]
        \centering
        \includegraphics[width=0.95\textwidth]{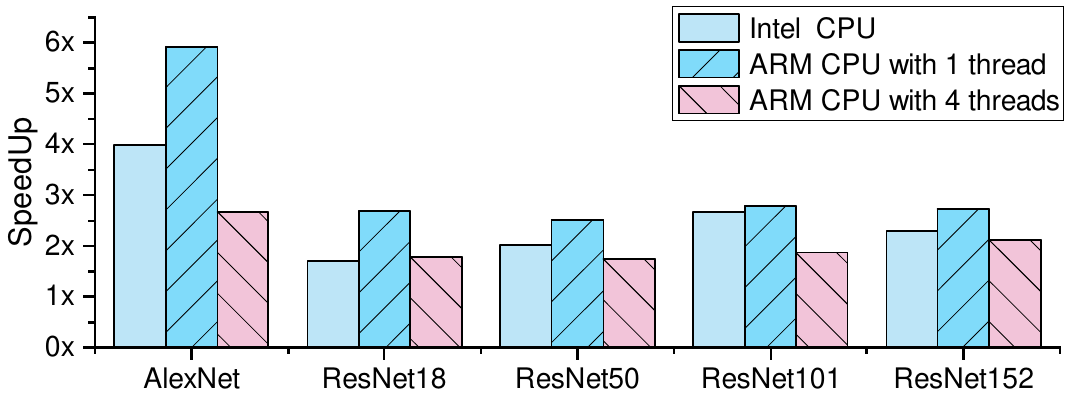}
        \caption{Speedup evaluation results on CPU. The height of the bar denotes the average acceleration rate of all selected epochs.
        }
        \label{fig-cpu-speedup}
    \end{figure}

    \paragraph{\textbf{Acceleration on desktop CPU.}}
    To examine the performance of our proposed approach in practical applications, we implement experiments on low computation power scenarios, where there exists an urgent need for acceleration in the training process.
    We use 1 core Intel CPU (Intel Xeon E5-2680 v4 2.4 GHz) as computation platform and Intel MKL as BLAS library for evaluation. We set $p$ = 99\% for ResNet-\{18,50,101,152\} and AlexNet on CIFAR-10 dataset and export the $\mathrm{d}\mathbf{O}$/$\mathbf{I}$/$\mathbf{W}$ from the training process of accuracy evaluation experiment every 50 epochs, and use those data to collect the latency of \textit{AFBP} and \textit{WGC} in our framework. The baseline of this experiment is the original back-propagation implementation of Caffe. According to the results in Fig. \ref{fig-cpu-speedup}, our algorithm can achieve $1.71\times$ $\sim$ $3.99\times$ speedup on average. These speedups refer to the acceleration of back-propagation while the forward stage is not included.

    \paragraph{\textbf{Acceleration on ARM CPU.}} We further evaluate our approach on ARM platform which is wildly used in edge computing.
    We choose Raspberry Pi 4B (with ARMv7 1500Hz) as experimental device and Eigen3 \cite{eigenweb} as BLAS library in this experiment because Intel MKL can't be deployed on ARM.
    In ARM experiment, we use the same setting as the desktop CPU experiment in section 6.3. Besides, we evaluate our approach with both single thread and four threads.
    According to Fig. \ref{fig-cpu-speedup}, in single thread experiments, the speedup of \texttt{Conv} Layer's back-propagation stage on ARM platform can be up to 5.92$\times$ with AlexNet. As for those networks that use \texttt{Conv-BN-RELU} as the basic module such as  ResNet-\{18,50,101,152\}, our approach can also achieve 2.52$\times$ $\sim$ 2.79$\times$ acceleration.
    On the other hand, the acceleration rate decrease in four thread experiment but can still reach 1.79$\times$ $\sim$ 2.78$\times$ speedup.
    The results illustrate that our algorithm still performs well on embedded device which is more urgent in reducing calculation time.

    \paragraph{\textbf{Comparison with existing works.}}
    Meprop \cite{sun2017meprop} has only experiments on MLP. \cite{wei2017minimal} supplements the CNN evaluation on the basis of Meprop \cite{sun2017meprop}. However, their chosen networks are unrepresentative because they are too naive to be adopted in practical applications. Based on \cite{wei2017minimal}, MSBP \cite{zhang2019memorized} makes further improvements, which is comparing with our method as illustrated in Table. \ref{tab:compare}. Our proposed algorithm can achieve higher sparsity than MSBP while keeping a better accuracy than baseline and MSBP on CIFAR-10.
    More importantly, the experiment result also shows that our work is also well performed on ImageNet which is more challenging but has not been evaluated in existing works.

      \begin{table}[h]
      \centering
      \caption{Comparison with MSBP \cite{zhang2019memorized}. The network and dataset are ResNet-18 and CIFAR-10. The definition of \texttt{acc}\% and $\rho_{nnz}$ can be found in Table \ref{tab:Imagenet}. }
      \label{tab:compare}
      \begin{threeparttable}
        \begin{tabular}{p{1.7cm}<{\centering}|p{1.2cm}<{\centering}|p{1.2cm}<{\centering}|p{4cm}<{\centering}|p{3.5cm}<{\centering}}
          \specialrule{0.8pt}{0pt}{0pt}
          {Method} & \texttt{acc}\% &  $\rho_{nnz}$ & Acceleration on Intel CPU & Acceleration On ARM\\
          \hline
          Baseline&95.08&1&1$\times$&1$\times$ \\
          MSBP \cite{zhang2019memorized} & 94.92& 0.4 & \  $\backslash$ & \ $\backslash$ \\
          Ours& \textbf{95.18} & \textbf{0.16} & \textbf{1.71$\times$} & \textbf{2.70$\times$}\\
          \specialrule{0.8pt}{0pt}{0pt}
        \end{tabular}
      \end{threeparttable}
  \end{table}


\section{Conclusion}

In this paper, we propose a new dynamically gradients pruning algorithm for CNN training. Different from the existing works, we assume the activation gradients of CNN satisfy normal distribution and then estimate their variance according to their average absolute value. After that, we calculate the pruning threshold according to the variance and a preset parameter $p$. The gradients are pruned randomly if they are under the threshold. Evaluations on state-of-the-art models have confirmed that our gradients pruning approach could accelerate the back-propagation up to 3.99$\times$ on desktop CPU and 5.92$\times$ on ARM CPU with a negligible accuracy loss.

\bibliographystyle{splncs04}

\begin{thebibliography}{10}
\providecommand{\url}[1]{\texttt{#1}}
\providecommand{\urlprefix}{URL }
\providecommand{\doi}[1]{https://doi.org/#1}

\bibitem{goyal2017accurate}
Goyal, P., Doll{\'a}r, P., Girshick, R., Noordhuis, P., Wesolowski, L., Kyrola,
  A., Tulloch, A., Jia, Y., He, K.: Accurate, large minibatch sgd: Training
  imagenet in 1 hour. arXiv preprint arXiv:1706.02677  (2017)

\bibitem{you2018imagenet}
You, Y., Zhang, Z., Hsieh, C.J., Demmel, J., Keutzer, K.: Imagenet training in
  minutes. In: Proceedings of the 47th International Conference on Parallel
  Processing. p.~1. ACM (2018)

\bibitem{jia2018highly}
Jia, X., Song, S., He, W., Wang, Y., Rong, H., Zhou, F., Xie, L., Guo, Z.,
  Yang, Y., Yu, L., et~al.: Highly scalable deep learning training system with
  mixed-precision: Training imagenet in four minutes. arXiv preprint
  arXiv:1807.11205  (2018)

\bibitem{cheng2018recent}
Cheng, J., Wang, P.s., Li, G., Hu, Q.h., Lu, H.q.: Recent advances in efficient
  computation of deep convolutional neural networks. Frontiers of Information
  Technology \& Electronic Engineering  \textbf{19}(1),  64--77 (2018)

\bibitem{micikevicius2017mixed}
Micikevicius, P., Narang, S., Alben, J., Diamos, G., Elsen, E., Garcia, D.,
  Ginsburg, B., Houston, M., Kuchaiev, O., Venkatesh, G., et~al.: Mixed
  precision training. arXiv preprint arXiv:1710.03740  (2017)

\bibitem{krizhevsky2012imagenet}
Krizhevsky, A., Sutskever, I., Hinton, G.E.: {ImageNet} classification with
  deep convolutional neural networks. In: Proceedings of the Advances in Neural
  Information Processing Systems. pp. 1097--1105 (2012)

\bibitem{simonyan2014very}
Simonyan, K., Zisserman, A.: Very deep convolutional networks for large-scale
  image recognition. arXiv preprint arXiv:1409.1556  (2014)

\bibitem{he2016deep}
He, K., Zhang, X., Ren, S., Sun, J.: Deep residual learning for image
  recognition. In: Proceedings of the IEEE Conference on Computer Vision and
  Pattern Recognition. pp. 770--778 (2016)

\bibitem{han2015deep}
Han, S., Mao, H., Dally, W.J.: Deep compression: Compressing deep neural
  networks with pruning, trained quantization and {Huffman} coding. arXiv
  preprint arXiv:1510.00149  (2015)

\bibitem{mao2017exploring}
Mao, H., Han, S., Pool, J., Li, W., Liu, X., Wang, Y., Dally, W.J.: Exploring
  the regularity of sparse structure in convolutional neural networks. arXiv
  preprint arXiv:1705.08922  (2017)

\bibitem{anwar2017structured}
Anwar, S., Hwang, K., Sung, W.: Structured pruning of deep convolutional neural
  networks. ACM Journal on Emerging Technologies in Computing Systems
  \textbf{13}(3), ~32 (2017)

\bibitem{lebedev2016fast}
Lebedev, V., Lempitsky, V.: Fast {ConvNets} using group-wise brain damage. In:
  Proceedings of the IEEE Conference on Computer Vision and Pattern
  Recognition. pp. 2554--2564 (2016)

\bibitem{luo2017thinet}
Luo, J.H., Wu, J., Lin, W.: Thinet: A filter level pruning method for deep
  neural network compression. In: Proceedings of the IEEE International
  Conference on Computer Vision. pp. 5058--5066 (2017)

\bibitem{he2017channel}
He, Y., Zhang, X., Sun, J.: Channel pruning for accelerating very deep neural
  networks. In: Proceedings of the IEEE International Conference on Computer
  Vision. pp. 1389--1397 (2017)

\bibitem{liu2017learning}
Liu, Z., Li, J., Shen, Z., Huang, G., Yan, S., Zhang, C.: Learning efficient
  convolutional networks through network slimming. In: Proceedings of the IEEE
  International Conference on Computer Vision. pp. 2736--2744 (2017)

\bibitem{wen2017coordinating}
Wen, W., Xu, C., Wu, C., Wang, Y., Chen, Y., Li, H.: Coordinating filters for
  faster deep neural networks. In: Proceedings of the IEEE International
  Conference on Computer Vision. pp. 658--666 (2017)

\bibitem{wen2017learning}
Wen, W., He, Y., Rajbhandari, S., Zhang, M., Wang, W., Liu, F., Hu, B., Chen,
  Y., Li, H.: Learning intrinsic sparse structures within long short-term
  memory. arXiv preprint arXiv:1709.05027  (2017)

\bibitem{aji2017sparse}
Aji, A.F., Heafield, K.: Sparse communication for distributed gradient descent.
  arXiv preprint arXiv:1704.05021  (2017)

\bibitem{prakash2018repr}
Prakash, A., Storer, J., Florencio, D., Zhang, C.: Repr: Improved training of
  convolutional filters. arXiv preprint arXiv:1811.07275  (2018)

\bibitem{sun2017meprop}
Sun, X., Ren, X., Ma, S., Wang, H.: meprop: Sparsified back propagation for
  accelerated deep learning with reduced overfitting. In: Proceedings of the
  34th International Conference on Machine Learning-Volume 70. pp. 3299--3308
  (2017)

\bibitem{wei2017minimal}
Wei, B., Sun, X., Ren, X., Xu, J.: Minimal effort back propagation for
  convolutional neural networks. arXiv preprint arXiv:1709.05804  (2017)

\bibitem{zhang2019memorized}
Zhang, Z., Yang, P., Ren, X., Sun, X.: Memorized sparse backpropagation. arXiv
  preprint arXiv:1905.10194  (2019)

\bibitem{gupta2015deep}
Gupta, S., Agrawal, A., Gopalakrishnan, K., Narayanan, P.: Deep learning with
  limited numerical precision. In: Proceedings of the International Conference
  on Machine Learning. pp. 1737--1746 (2015)

\bibitem{zhou2016dorefa}
Zhou, S., Wu, Y., Ni, Z., Zhou, X., Wen, H., Zou, Y.: {DoReFa-Net}: Training
  low bitwidth convolutional neural networks with low bitwidth gradients. arXiv
  preprint arXiv:1606.06160  (2016)

\bibitem{park2018value}
Park, E., Yoo, S., Vajda, P.: Value-aware quantization for training and
  inference of neural networks. In: Proceedings of the European Conference on
  Computer Vision. pp. 580--595 (2018)

\bibitem{szegedy2016rethinking}
Szegedy, C., Vanhoucke, V., Ioffe, S., Shlens, J., Wojna, Z.: Rethinking the
  inception architecture for computer vision. In: Proceedings of the IEEE
  conference on computer vision and pattern recognition. pp. 2818--2826 (2016)

\bibitem{wen2017terngrad}
Wen, W., Xu, C., Yan, F., Wu, C., Wang, Y., Chen, Y., Li, H.: {TernGrad}:
  Ternary gradients to reduce communication in distributed deep learning. In:
  Proceedings of the Advances in Neural Information Processing Systems. pp.
  1509--1519 (2017)

\bibitem{bottou1998online}
Bottou, L.: Online learning and stochastic approximations. On-Line Learning in
  Neural Networks  \textbf{17}(9), ~142 (1998)

\bibitem{paszke2017automatic}
Paszke, A., Gross, S., Chintala, S., Chanan, G., Yang, E., DeVito, Z., Lin, Z.,
  Desmaison, A., Antiga, L., Lerer, A.: Automatic differentiation in {PyTorch}.
  In: NIPS Workshop (2017)

\bibitem{jia2014caffe}
Jia, Y., Shelhamer, E., Donahue, J., Karayev, S., Long, J., Girshick, R.,
  Guadarrama, S., Darrell, T.: Caffe: Convolutional architecture for fast
  feature embedding. arXiv preprint arXiv:1408.5093  (2014)

\bibitem{krizhevsky2009learning}
Krizhevsky, A., Hinton, G.: Learning multiple layers of features from tiny
  images. Tech. rep., Citeseer (2009)

\bibitem{deng2009imagenet}
Deng, J., Dong, W., Socher, R., Li, L.J., Li, K., Li, F.F.: {ImageNet}: A
  large-scale hierarchical image database. In: Proceedings of the IEEE
  Conference on Computer Vision and Pattern Recognition. pp. 248--255 (2009)

\bibitem{eigenweb}
Guennebaud, G., Jacob, B., et~al.: Eigen v3. http://eigen.tuxfamily.org (2010)

\end{thebibliography}

\end{document}